\documentclass{article}

\PassOptionsToPackage{numbers,compress}{natbib}
\usepackage[preprint]{neurips_2026}

\usepackage{url}
\usepackage{booktabs}
\usepackage{amsfonts}
\usepackage{amssymb}
\usepackage{amsmath}
\usepackage{nicefrac}
\usepackage{microtype}
\usepackage[table]{xcolor}
\usepackage{algorithm}
\usepackage{algorithmic}
\usepackage{enumitem}
\usepackage{graphicx}
\usepackage{adjustbox}
\usepackage{subcaption}
\usepackage{wrapfig}
\usepackage{multirow}
\usepackage{makecell}
\usepackage{colortbl}
\usepackage{array}
\usepackage{hyperref}
\usepackage{xspace}

\definecolor{tabgray}{gray}{0.92}

\newcommand{\method}{FVAttn\xspace}
\newcommand{\rlb}{RLB\xspace}
\newcommand{\sasa}{SASA\xspace}

\title{FVAttn: Adaptive Sparse Attention with Runtime Load Balancing for Video Generation}

\hypersetup{
  pdftitle={FVAttn: Adaptive Sparse Attention with Runtime Load Balancing for Video Generation},
  pdfauthor={Hao Liu},
  pdfkeywords={video diffusion, sparse attention, distributed inference, runtime load balancing, algorithm-system co-design},
}

\author{
Hao Liu$^{1,2}$ \quad Chenghuan Huang$^{2}$ \quad Ye Huang$^{4}$ \quad Zhiying Wen$^{1}$ \quad Hao Liu$^{3}$\\
Mohan Zhang$^{3}$ \quad Chen Li$^{3}$ \quad Ziyang Ma$^{2}$ \quad Jing Lyu$^{3}$ \quad Jiangsu Du$^{1}$ \\[4pt]
$^{1}$Sun Yat-sen University \quad
$^{2}$WeChat HPC, Tencent Inc. \\
$^{3}$WeChat Vision, Tencent Inc. \quad
$^{4}$Peking University \\
\texttt{\{liuh393,wenzhy35\}@mail2.sysu.edu.cn} \quad \texttt{\{dujiangsu\}@mail.sysu.edu.cn} \\
\texttt{\{huangye26\}@stu.pku.edu.cn} \\
\texttt{\{bighhliu,caderhuang,leweshaoliu,zeromhzhang,chaselli,ziyangma,eckolv\}@tencent.com}
}

\begin{document}
\maketitle

\begin{abstract}

Video Diffusion Transformers process long spatio-temporal sequences, making self-attention the main bottleneck in high-resolution video generation.
Training-free sparse attention reduces this cost, but adaptive Top-$p$ routing creates uneven per-head workloads under multi-GPU sequence parallelism.
The resulting workload heterogeneity turns sparse attention into a rank-level straggler problem. 
We present \method{}, a training-free sparse-attention system that improves the distributed execution efficiency of adaptive sparse attention under multi-GPU sequence parallelism.
\method{} uses Top-$p$ routing, a Top-$k$ safety floor, and video-aware block organization as the sparse-routing frontend, then repairs the materialized mask at runtime.
Runtime Load Balancing migrates a small number of heavy heads via P2P communication to shorten the current critical path.
Slack-Aware Sparse Augmentation fills residual non-critical-rank slack with additional high-value blocks, while overlap hides scheduling and migration overhead behind existing computation.
On step-distilled Wan2.2 I2V, \method{} reduces average load imbalance from 1.34 to 1.08 and delivers a $4.41\times$ attention speedup over FlashAttention, while achieving a $2.02$--$2.11\times$ DiT inference speedup with competitive video quality.

\end{abstract}

\section{Introduction}

Recently, Video Diffusion Transformers (Video DiTs)~\cite{wan2025wan,wu2025hunyuanvideo,yang2025cogvideox,zheng2024open,bao2024viduhighlyconsistentdynamic} have achieved remarkable progress in visual quality. However, their inference cost remains prohibitively high. As video resolution, frame count, and duration continue to increase, Video DiTs must process rapidly growing spatio-temporal token sequences, leading to a substantial increase in inference latency.
For example, generating a 5-second 720p video with Wan2.2-14B I2V on a single NVIDIA H20 GPU still takes approximately two hours, with attention accounting for 74.1\% of the total inference time.
To improve deployment efficiency, step-distillation~\cite{yin2024improved,yin2024one,yin2025slow} has been widely adopted in industry, which reduces the number of denoising steps and can provide up to a $25\times$ inference speedup. 
However, step distillation leaves the per-step sequence length and the resulting quadratic attention cost unchanged. Attention therefore remains a major bottleneck even in the few-step regime, motivating complementary optimizations within each denoising step.

Sparse attention provides such a complementary optimization by exploiting the substantial redundancy in video attention~\citep{zhang2025spargeattn,xi2025sparse,yang2026sparse,zhang2026training, zhang2025sla}.
For a query token or query block, only a small subset of key blocks contributes substantially to the output.
Training-free sparse-attention methods exploit this property by constructing block-level masks at inference time and skipping low-value query-key interactions.
Among these methods, Top-$p$ routing is attractive as it adapts the retained-block count to the concentration of each attention distribution.
A concentrated head can keep only a few high-mass blocks, while a flat or multi-modal head can retain more interactions.
Compared with a fixed Top-$k$ budget, Top-$p$ can better preserve attention fidelity under a similar average compute budget.

However, this adaptivity introduces a new systems bottleneck.
Top-$p$ routing yields varying numbers of retained blocks across attention heads and query regions, resulting in imbalanced workloads under Ulysses-style sequence parallelism~\citep{jacobs2023deepspeed}, where a sequence-to-head All-to-All operation distributes complete-sequence attention heads across GPU ranks. 
When multiple high-density heads are mapped to the same rank, the resulting workload skew causes stragglers during the synchronized attention phase, forcing other ranks to remain idle after completing their assigned computation. 
Consequently, the theoretical efficiency gains from sparsity are partially offset by distributed execution overhead. 
This motivates the key question addressed in this paper: \emph{how can we retain the fidelity advantages of adaptive sparse attention while achieving balanced and efficient distributed execution?}

\begin{wrapfigure}{r}{0.50\linewidth}
  \centering
  \includegraphics[width=\linewidth]{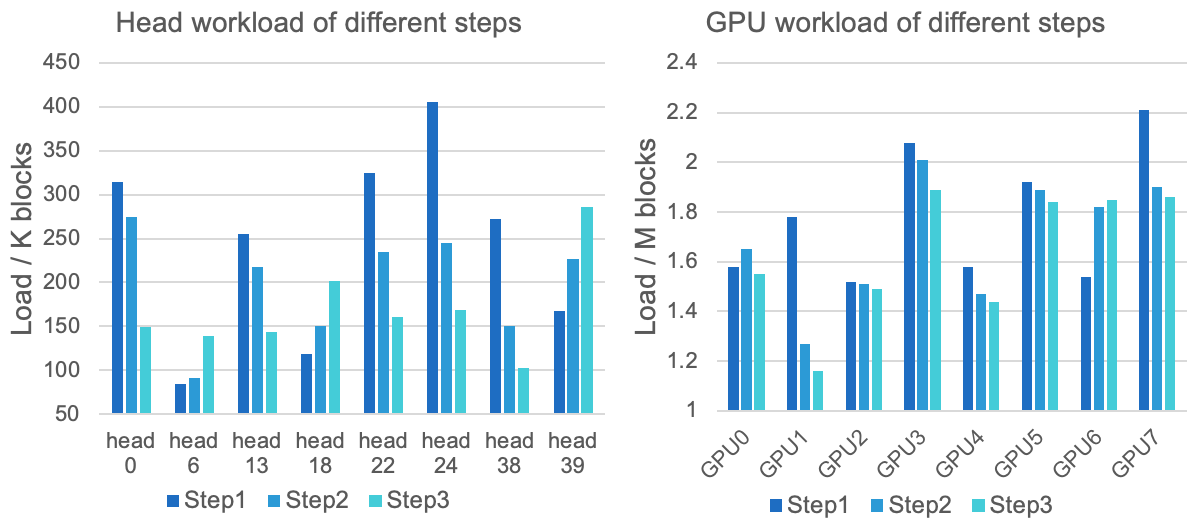}
  \caption{Step-to-step workload variation in few-step video generation, with maximum adjacent-step changes of 97\% at the head level and 44\% at the rank level.}
  \label{fig:crossstep}
  \vspace{-12pt}
\end{wrapfigure}

Existing sparse sequence-parallel systems often try to choose head or block layouts before the current attention computation using historical sparsity, offline profiling, or approximate similarity signals~\citep{chen2025db}.
This strategy is fragile in few-step video generation.
As shown in Figure~\ref{fig:crossstep}, the maximum adjacent-step variation reaches $97\%$ for the same head density and $44\%$ for the same rank load in our 4-step distilled setting.
More fundamentally, under Ulysses-style sequence parallelism, the true workload of the current sparse-attention layer becomes observable only after the sequence-to-head All-to-All completes and the current sparse mask is materialized.
A pre-layout policy must therefore rely on stale or approximate signals.
A full post-mask repartition could use the true workload, but it would require another large data movement and can erase the sparse-attention gain.

\begin{figure}[tbp]
  \centering
  \includegraphics[width=0.88\linewidth]{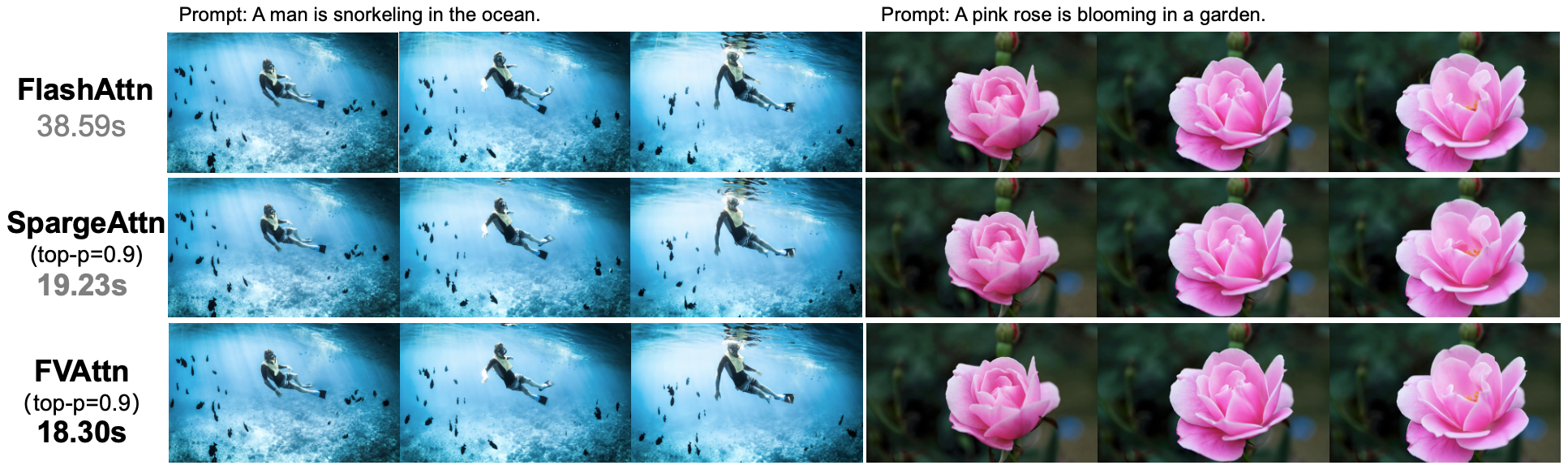}
  \caption{\method{} combines efficient sparse attention with runtime load balancing to accelerate video generation while preserving visual quality.}
  \label{fig:visual_result}
  \vspace{-12pt}
\end{figure}

Our key observation is that the load imbalance introduced by Top-$p$ routing is highly localized: the majority of the workload skew is typically concentrated in only a small subset of heavy heads.
This observation indicates that eliminating attention stragglers does not require expensive global repartitioning or redesigning the entire parallel layout. 
Instead, by identifying and migrating only a small number of overloaded heads after the actual sparse pattern is materialized, we can recover most of the critical-path efficiency while introducing only bounded communication overhead. 
Motivated by this insight, \method{} introduces Runtime Load Balancing (\rlb{}), a lightweight mechanism that adapts the distributed execution layout to the realized workload. 
After sparse-mask materialization, \rlb{} profiles the actual per-head workload and migrates selected heads through efficient peer-to-peer communication. 
In the default 8-GPU setting, each rank sends and receives at most one local head, corresponding to only approximately $20\%$ of its assigned head set. 
Therefore, rather than predicting an optimal future partitioning, \rlb{} directly repairs the imbalance of the current attention invocation based on the observed workload.

Reducing the straggling rank alone is insufficient to fully exploit the available computational capacity.
After bounded migration, some non-critical ranks may still have residual slack under the global critical path.
\method{} converts this slack into useful computation through Slack-Aware Sparse Augmentation.
Instead of globally increasing the Top-$p$ threshold, which would increase every rank and extend the makespan, \sasa{} adds high-value blocks only on ranks with measured slack.
This turns synchronization idle time into additional sparse-mask coverage while preserving the global attention latency.

\method{} begins with an adaptive sparse-routing stage that constructs the initial mask and produces two forms of runtime metadata: per-head workloads and block-importance rankings. 
\rlb{} uses the former to redistribute selected heads across ranks without changing the sparse computation itself, while \sasa{} uses the latter to spend residual slack on additional high-value blocks without extending the attention makespan. 
Thus, routing decides what to compute, \rlb{} decides where to compute it, and \sasa{} converts otherwise idle capacity into improved sparse-mask coverage.

We evaluate \method{} on Wan2.2 I2V, Wan2.2 Animate, and Wan2.1 T2V under 4-step distilled inference~\citep{lightx2v}. 
On Wan2.2 I2V, \rlb{} reduces the average load-imbalance factor from $1.34$ to $1.08$, while the complete \method{} stack achieves a $4.41\times$ attention speedup over FlashAttention with only $0.7$\,ms of visible runtime overhead. 
Across the evaluated workloads, \method{} consistently improves the quality--efficiency Pareto frontier over existing training-free sparse-attention. Our contributions are as follows.
\begin{itemize}[leftmargin=*]
  \item We identify a runtime systems bottleneck in adaptive sparse attention for distributed video DiT inference: fidelity-improving Top-$p$ routing induces head-level workload heterogeneity that becomes rank-level stragglers under sequence parallelism.
  \item We propose \rlb{}, a post-routing runtime repair mechanism that uses realized per-head workloads after sparse-mask materialization and performs communication-budgeted P2P head migration to shorten the current attention critical path.
  \item We propose \sasa{}, which converts residual non-critical-rank slack into additional high-value sparse blocks, improving sparse-mask coverage without globally increasing the Top-$p$ threshold or extending the attention makespan.
  \item We implement \method{} as an overlapped sparse-attention runtime and show improvements in load balance, attention latency, DiT latency, and video quality metrics on multiple video DiT workloads.
\end{itemize}

\section{Background and Related Work}
\label{sec:background}

\subsection{Training-Free Sparse Attention for Video DiTs}

Training-free sparse attention for video DiTs can be broadly categorized into static sparse patterns and dynamic sparse routing. 
Static methods~\cite{zhang2025fast,li2026radial} use predetermined or content-agnostic structures, whereas dynamic methods~\cite{zhang2025spargeattn,xi2025sparse,yang2026sparse,zhang2026training} construct input-dependent masks from the current attention features. 
\method{} adopts dynamic block-sparse attention. For attention head $h$, query block $i$, and key block $j$, a routing module estimates an importance score $s_{h,i,j}$ and generates a binary mask $M_{h,i,j}\in\{0,1\}$, where only selected query--key block pairs are evaluated.

Sparse-routing policies differ primarily in how they allocate computation. Top-$k$ retains the $k$ highest-scoring key blocks for every query block, yielding a fixed and relatively regular workload. Top-$p$ instead retains the smallest set of blocks whose cumulative normalized importance reaches threshold $p$. It therefore assigns more computation to flat or multi-peaked attention distributions and less to concentrated ones, improving adaptability but producing heterogeneous workloads across query blocks and heads.

\method{} combines Top-$p$ routing with a Top-$k$ safety floor. Top-$p$ provides an adaptive computation budget, while the Top-$k$ floor maintains a minimum budget against routing-estimation errors or overly aggressive sparsification. This frontend preserves attention fidelity but introduces heterogeneous per-head workloads. Under fixed Ulysses head placement, these workload differences directly translate into rank-level runtime imbalance.

\subsection{Sequence Parallelism and Load Imbalance}

Long-sequence video DiT inference commonly relies on sequence parallelism, including Ring Attention~\citep{liu2024ringattention}, Ulysses~\citep{jacobs2023deepspeed}, and unified sequence-parallel strategies such as USP~\citep{fang2024usp}.
This work is mainly built on Ulysses-style sequence parallelism.
A sequence-to-head All-to-All converts sequence shards into head shards, so each GPU holds complete sequences for a subset of attention heads and computes local attention independently.
Dense attention is naturally balanced in this layout because all heads have nearly identical compute.
Dynamic sparse attention breaks this property because each head's actual compute is determined by the current sparse mask.

Specifically, the realized workload of head $h$ and the workload of rank $r$ with local head set $\mathcal{H}_r$ are defined as
\begin{equation}
L_h = \operatorname{mean}_{i}\sum_j M_{h,i,j}, \qquad L_r = \sum_{h\in\mathcal{H}_r} L_h.
\end{equation}
Because the synchronized attention phase waits for all ranks, the latency is governed by $\max_r L_r$.
We use the load-imbalance factor
\begin{equation}
\rho = \frac{\max_r L_r}{\frac{1}{R}\sum_{r=1}^{R} L_r},
\end{equation}
where $R$ is the number of ranks.
A value close to $1$ indicates balanced execution.
A much larger value means that many GPUs waste time waiting for the slowest rank.
The later reported reduction from $1.34$ to $1.08$ refers to this metric.

Existing distributed sparse-attention systems mitigate workload imbalance through mask-aware repartitioning, profile-based head placement, or pattern-specific parallel execution~\citep{chen2025db,liu2026s,lidsa,ge2026osp}. For example, db-SP globally repartitions sparse workloads at the head and block levels and amortizes planning through cross-step reuse, while S-HPLB derives head placement from offline-profiled sparsity characteristics. DSA and OSP-Next instead exploit structured sparse patterns to construct specialized parallel execution layouts. In contrast, \method{} preserves the existing Ulysses layout, waits until the current adaptive mask is materialized, and performs communication-budgeted P2P repair directly from its realized per-head workloads.

\section{FVAttn Design}
\label{sec:method}

\subsection{Overview}

\method{} starts from a systems bottleneck of dynamic sparse attention in multi-GPU inference.
Top-$p$ routing improves mask fidelity, but its head-level sparsity differences become rank-level realized workload imbalance under Ulysses-style sequence parallelism and slow down the synchronized attention phase.
Therefore, \method{} performs two-stage runtime scheduling after the current sparse mask is materialized.
It first uses \rlb{} to repair the critical path according to the realized per-head workload.
It then uses \sasa{} to convert residual slack on non-critical ranks into additional high-value blocks.

\begin{figure}[tbp]
  \centering
  \includegraphics[width=0.88\linewidth]{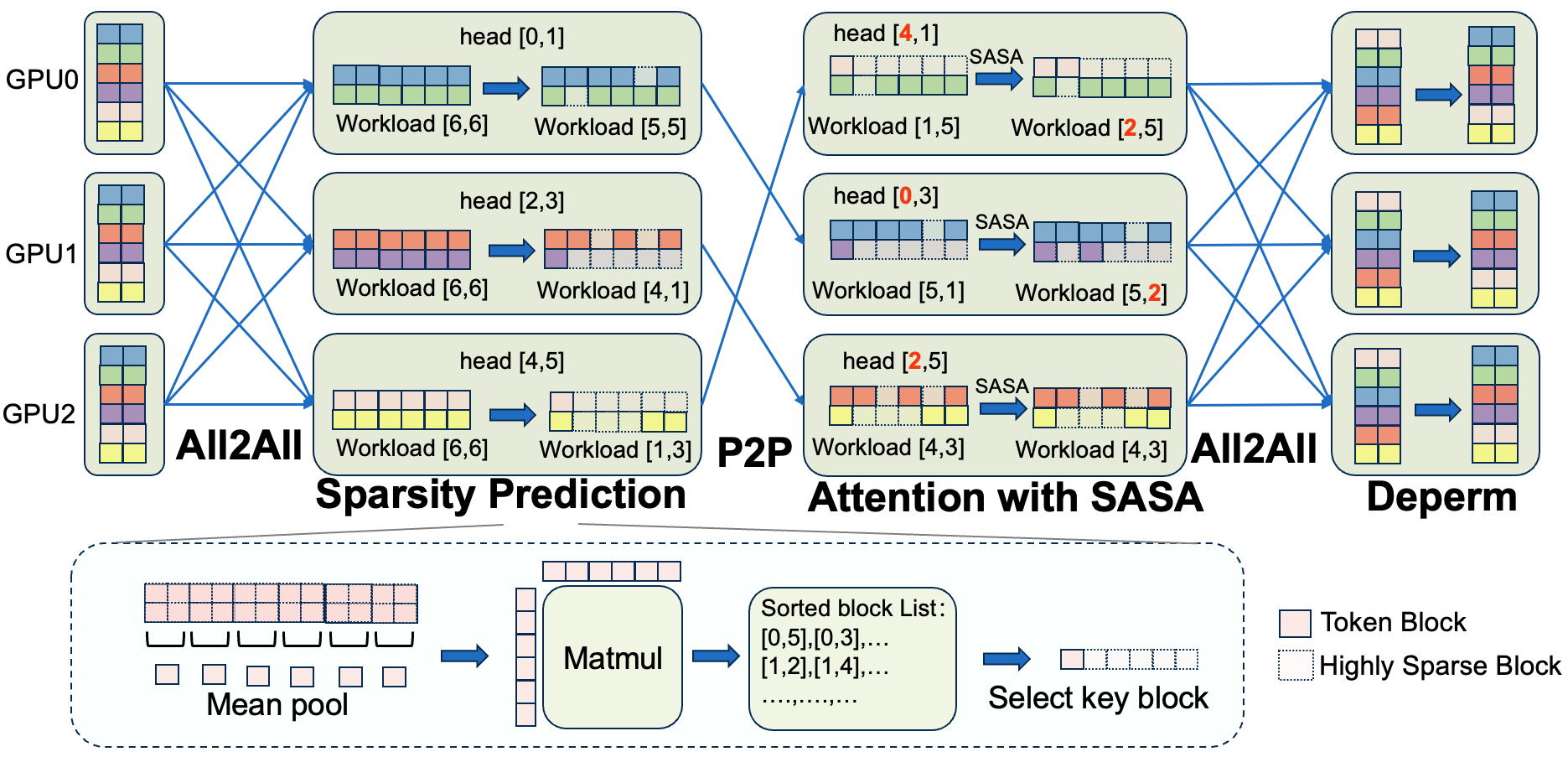}
  \caption{Overview of \method{} after sequence-to-head All-to-All, including mask routing, \rlb{} with P2P head migration for load-imbalance reduction, slack-aware sparse augmentation for quality improvement, and block-sparse attention.}
  \label{fig:overview}
\end{figure}

Concretely, after sequence-to-head All-to-All, each sparse attention layer executes the following pipeline.
First, \method{} constructs the adaptive sparse mask.
Then, \method{} measures the realized per-head workload and repairs imbalance with constrained P2P communication.
Next, \method{} converts the residual rank-local slack after balancing into additional mask budget.
Then, \method{} runs block-sparse attention.
Finally, \method{} performs reverse All-to-All and restores the head order.
Sections~\ref{sec:rlb} and~\ref{sec:sasa} introduce the two core runtime mechanisms of \method{}.
\rlb{} repairs the realized workload imbalance caused by Top-$p$ routing and shortens the critical path of the synchronized attention phase.
\sasa{} converts the residual slack exposed after load balancing into additional high-value sparse-block computation, improving sparse-mask coverage and video quality without extending the critical path.

In the above pipeline, the sparse-routing frontend is the shared input to \rlb{} and \sasa{}.
To improve routing-stage similarity estimation, \method{} reorders blocks along a Hilbert curve so that spatially adjacent tokens tend to fall into nearby blocks across scales.
On block-level similarities from pooled Q/K, \method{} mainly uses a Top-$p$ CDF threshold to adaptively determine the retained-block count according to the distribution shape of each query block.
\method{} also uses a Top-$k$ floor as a safety net against over-sparsification under extremely concentrated distributions.
The same routing pass also produces the per-head density required by \rlb{} and the key-block importance order required by \sasa{}.
Therefore, the two runtime mechanisms require no additional routing computation.
Top-$p$ adaptivity is thus the source of mask-fidelity gains and the workload heterogeneity that \rlb{} must handle in Section~\ref{sec:rlb}.

\subsection{Runtime Load Balancing}
\label{sec:rlb}

Once the dynamic sparse mask is constructed, the realized workload $L_h$ of each head is fully observable.
The workload of rank $r$ is then fixed as $L_r=\sum_{h\in\mathcal{H}_r}L_h$.
The adaptive budget of Top-$p$ routing makes different heads retain different numbers of key blocks, creating head-level variation in $L_h$.
This variation is further amplified into rank-level load imbalance, often with $\rho\gg1$.
Because the synchronized attention phase must wait for the slowest rank, its completion time is approximately governed by $\max_r L_r$.
An overloaded rank becomes a straggler, while the other ranks produce idle bubbles that partially offset the algorithmic compute savings from Top-$p$ routing.

To mitigate these stragglers and idle bubbles, \method{} redistributes work across ranks using heads as the scheduling unit.
We formulate this load-balancing procedure as an explicit benefit optimization.
A schedule reassigns heads across ranks and yields post-balancing workloads $L_r^{\text{new}}$.
Using $c$ as the unit compute time of one sparse block, the critical-path benefit of the schedule is
\begin{equation}
K = c\Big(\max_r L_r - \max_r L_r^{\text{new}}\Big).
\end{equation}
The cost consists of the communication overhead $P$ for moving head states and the scheduling overhead $C$ for solving the migration plan.
The net gain is
\begin{equation}
G = K - P - C.
\end{equation}
\begin{wrapfigure}{r}{0.40\linewidth}
  \vspace{-0.8em}
  \centering
  \includegraphics[width=\linewidth]{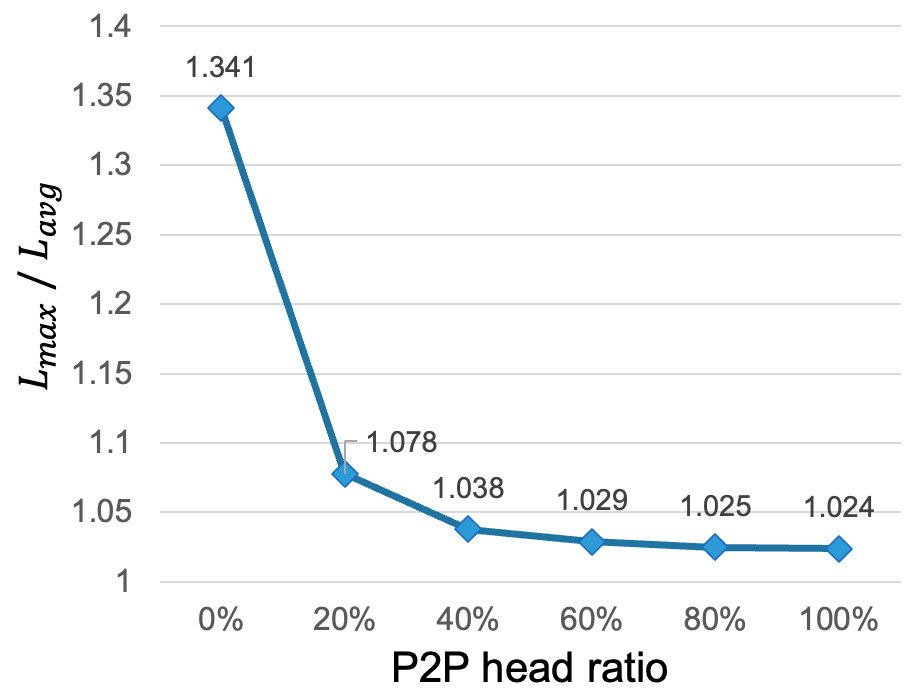}
  \caption{Load-imbalance factor under different RLB head-migration budgets.}
  \label{fig:migration-budget}
  \vspace{-0.8em}
\end{wrapfigure}
This decomposition explains why \method{} must trade off additional overhead against load-balancing benefit, rather than simply maximizing $K$.
Predictive pre-scheduling before sequence-to-head All-to-All can fold communication into the existing collective, so $P\approx0$.
However, it relies on predicted rather than current workloads, and $L_h$ can vary sharply across adjacent steps in step-distilled video DiTs.
As a result, the realized $K$ can shrink substantially, and can even become negative.
Full post-mask repartition after mask materialization can drive the load close to perfect balance and make $K$ approach its upper bound.
However, it requires another large-scale collective data movement, making $P$ too large.
Both alternatives therefore make $G$ small.
\method{} takes a third path: it searches for a high-gain schedule inside a constrained P2P head-migration space.

We observe that the load skew introduced by Top-$p$ is concentrated in a small number of heads, rather than uniformly spread across all heads.
This means that moving a few dense heads from overloaded ranks to ranks with remaining capacity can remove most of the straggler time without the communication cost of global reshuffling.
On Wan2.2 I2V, migrating only $20\%$ of local heads, which is one head per rank in the 8-GPU setting, reduces average imbalance from $1.34$ to $1.08$.
Increasing the migration budget further yields much smaller gains.
Based on this observation, \method{} restricts the candidate schedule space from global repartitioning to lightweight P2P head migration.
In the default configuration, this constraint means that each rank migrates at most one local head, or roughly $20\%$ of its local heads.
The migration plan can then be represented as a permutation $\sigma$ over ranks, where fixed points $\sigma(r)=r$ denote no migration.
Rank $r$ sends a local head $h_r$ to $\sigma(r)$ and receives one head from $\sigma^{-1}(r)$.
The post-balancing workload is
\begin{equation}
L_r^{\text{new}} = L_r - L_{h_r} + L_{h_{\sigma^{-1}(r)}}.
\end{equation}
This constraint does not guarantee globally optimal load balance.
It deliberately gives up the full feasible space of arbitrary subset redistribution in exchange for tiny single-head P2P traffic and low matching overhead.
Within this constrained feasible set, \rlb{} jointly searches migration tuples $(h,r_{\mathrm{src}},r_{\mathrm{dst}})$, minimizing $L_{\max}^{\mathrm{new}}$ as the primary objective and rank-load variance as the tie-breaker:
\begin{equation}
\sigma^\star = \arg\min_{\sigma \in \mathcal{S}_{20\%}}\ \big(\max_r L_r^{\text{new}},\ \operatorname{Var}_r(L_r^{\text{new}})\big),
\end{equation}
where $\mathcal{S}_{20\%}$ denotes candidate schedules that satisfy the one-send-one-receive constraint and migrate no more than roughly $20\%$ of local heads.
In other words, \rlb{} does not pursue the theoretical optimum under an additional global reshuffle.
Instead, it minimizes the critical-path workload inside a candidate space where communication and search costs are controlled.

To reduce $C$, all ranks gather global density information through one lightweight all-gather.
Because this information is identical on all ranks and the constrained search is deterministic, each GPU independently runs the same search and extracts its own send and receive partners.
This avoids a rank-0 solve-and-broadcast synchronization.
The search strategy depends on the world size and communication constraint.
For the evaluated $R=8$ setting, \method{} can build pair topologies satisfying the one-send-one-receive constraint, enumerate candidate matchings in $\mathcal{S}_{20\%}$, and choose the best plan under the lexicographic objective above.
For more general world sizes, \method{} can also build a unidirectional ring topology so that every rank participates in one send and one receive.
These constrained searches produce deterministic schedules within hundreds of microseconds.
Thus, $C$ is small relative to $K$ and can be embedded directly into the synchronized attention forward path.

\subsection{Slack-Aware Sparse Augmentation}
\label{sec:sasa}

Runtime Load Balancing repairs the ``slowest-rank'' problem, but bounded whole-head migration and the one-send-one-receive constraint usually leave the ranks not perfectly balanced after \rlb{}.
Non-critical ranks may still finish local computation before the new slowest rank, and this waiting time forms the residual slack after \rlb{}.
Traditional sparse attention treats mask routing as a local decision independent of the inference environment.
Its sparsity is mainly determined by attention-score distributions and preset parameters, without considering whether a rank remains idle after \rlb{}.
For these waiting non-critical ranks, extra computation does not increase the overall latency as long as it stays within the residual slack.
Based on this observation, \method{} proposes \sasa{} to use residual slack after \rlb{} for computing more high-value attention blocks.

Concretely, \sasa{} formulates this process as resource redistribution under a critical-path constraint.
When entering the \sasa{} stage, the residual slack from non-critical rank $r$ to the current critical path is
\begin{equation}
\Delta L_r = L_{\max}^{\text{new}} - L_r^{\text{new}}, \qquad L_{\max}^{\text{new}}=\max_r L_r^{\text{new}}.
\end{equation}
Here, $L_r^{\text{new}}$ denotes the current workload when \sasa{} starts, which can be either the workload after \rlb{} or the original post-mask workload when \rlb{} is skipped.
For each rank with sufficiently large slack, \sasa{} does not fill the entire slack directly.
Instead, it allocates an extra sparse budget
\begin{equation}
B_r = n_1 \Delta L_r,
\end{equation}
where $n_1$ is the augmentation coefficient.
Then \sasa{} follows the importance order already produced by routing and uses $B_r$ to add extra key blocks into the mask, increasing the density of selected heads on that rank.
The augmented workload is
\begin{equation}
L_r^{\text{aug}} = L_r^{\text{new}} + B_r.
\end{equation}

Sparse-attention quality in video generation is strongly related to mask density.
Under Top-$p$ routing, retaining more key blocks covers a larger fraction of the original dense-attention mass and loses less information during sparse attention.
\sasa{} is not equivalent to globally increasing the Top-$p$ threshold.
A global threshold increase raises the workload of all ranks and can lengthen the critical path.
In contrast, \sasa{} only appends not-yet-selected high-value blocks on non-critical ranks that still have slack after load balancing.
Thus, it converts idle compute that would have been wasted by synchronization into higher sparse-mask coverage without increasing the overall latency.

This augmentation is not free by itself.
If the augmentation budget $B_r$ of a rank exceeds its residual slack $\Delta L_r$, $L_r^{\text{aug}}$ can overtake the previous maximum load $L_{\max}^{\text{new}}$ and create a new straggler.
In addition, candidate scanning and scheduling introduce a fixed overhead $S$.
The visible critical-path overhead of \sasa{} is therefore
\begin{equation}
A = \max\big(0,\ \max_r L_r^{\text{aug}} - L_{\max}^{\text{new}}\big) + S.
\end{equation}
\method{} aims to make $A\to0$ by satisfying two conditions.
First, each rank should keep its augmentation budget within its residual slack, i.e., $B_r\le\Delta L_r$, so that no new straggler is created.
Second, the scheduling overhead $S$ should be minimized or hidden.

The trigger threshold and the augmentation coefficient implement the first condition.
The trigger threshold $n_2$ starts augmentation only when $\Delta L_r > n_2 L_{\max}^{\text{new}}$, avoiding unnecessary scheduling work for tiny slack.
The augmentation coefficient $n_1$ discounts $\Delta L_r$ into a safe sparse budget $B_r=n_1\Delta L_r$, leaving margin for kernel granularity and estimation error and preventing $B_r$ from effectively overtaking $\Delta L_r$.
In our 8$\times$H20, world-size-8 setting, we set $n_2=0.07$ and $n_1=0.8$.
These values can be recalibrated for hardware with different kernel-launch overheads or communication-to-computation ratios.
The second condition is met by reusing routing outputs and overlapping scheduling.
Additional blocks are selected from the already sorted key-block order, excluding heads migrated out by the current \rlb{} step, and the mask is updated in place.
No extra sorting or QK scoring is required.
The augmentation work is scheduled after P2P migration is issued and before synchronization, so much of its overhead is hidden by the communication window.

\begin{figure}[tbp]
  \centering
  \includegraphics[width=1.0\linewidth]{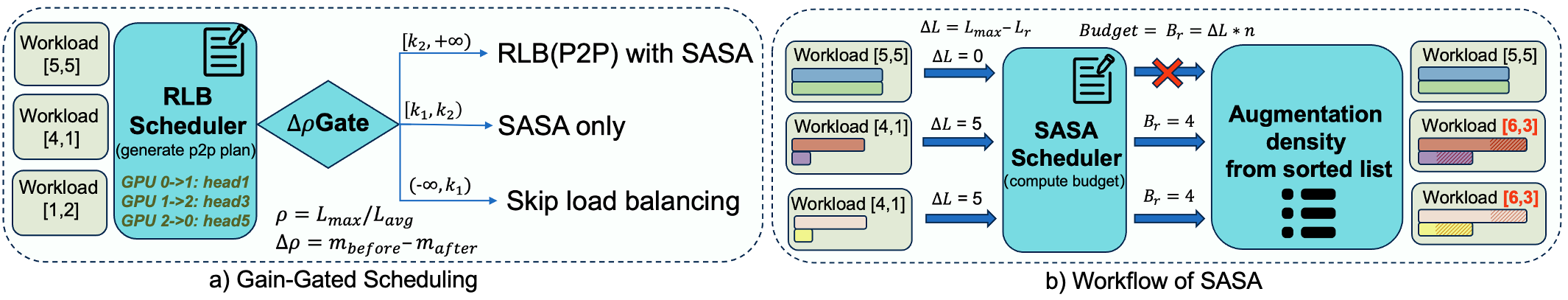}
  \caption{(a) Runtime mechanisms are enabled according to the observable load imbalance. (b) The \sasa{} workflow.}
  \label{fig:adaptive-compute}
\end{figure}

The two mechanisms are enabled according to a cost-benefit trade-off.
After the global workload is known, if the load-imbalance factor satisfies $\rho<k_1$, the workload is close to balanced and the system skips extra runtime mechanisms.
If $k_1\le\rho<k_2$, the imbalance has exposed some slack on non-critical ranks but is not large enough to offset the communication and scheduling cost of P2P migration.
In this case, \method{} enables only the lower-overhead \sasa{} and uses the slack in the current post-mask workloads to augment high-value blocks.
If $\rho\ge k_2$, the imbalance is severe enough that \method{} first enables \rlb{} to repair the critical path through P2P head migration, and then applies \sasa{} on the residual slack after \rlb{}.
Here $k_2\ge k_1$.
This piecewise behavior reflects the overall principle of \method{}: skip extra mechanisms under low imbalance, enable only low-overhead \sasa{} under medium imbalance, and enable \rlb{}+\sasa{} under high imbalance, so that runtime mechanisms take effect progressively according to observable load state.

\subsection{Efficient Implementation}
\label{sec:overlap}

With \sasa{}, augmentation introduces an additional visible critical-path cost $A$ beyond the \rlb{} costs in Section~\ref{sec:rlb}.
The joint net gain becomes
\begin{equation}
G = K - P - C - A,
\end{equation}
where $P$ is P2P communication, $C$ is runtime scheduling, and $A$ is slack-aware augmentation overhead.
The algorithmic objective is to keep $K$ large; the systems objective is to make $P$, $C$, and $A$ mostly invisible.

\method{} uses two asynchronous overlap paths.
First, it hides the visible part of $C$ with CPU--GPU overlap: the CPU searches the \rlb{} plan and computes part of the \sasa{} budget while the GPU performs mandatory V quantization.
Since these tasks are independent, scheduling and budget computation are largely absorbed by the quantization window.

Second, it hides the visible parts of $P$ and $A$ with computation--communication overlap.
After the balancing plan is ready, P2P head migration is issued asynchronously; the runtime then proceeds with V quantization for non-migrated heads and \sasa{} augmentation, and synchronizes only when migrated states are consumed.
Thus, communication uses network resources while useful computation occupies the GPU.
For the flash fallback branch, where the migration payload is larger and no \sasa{} work is available, \method{} instead splits attention into local-head and migrated-head computation and overlaps the local-head path with P2P transfer.

\begin{figure}[tbp]
  \centering
  \includegraphics[width=0.82\linewidth]{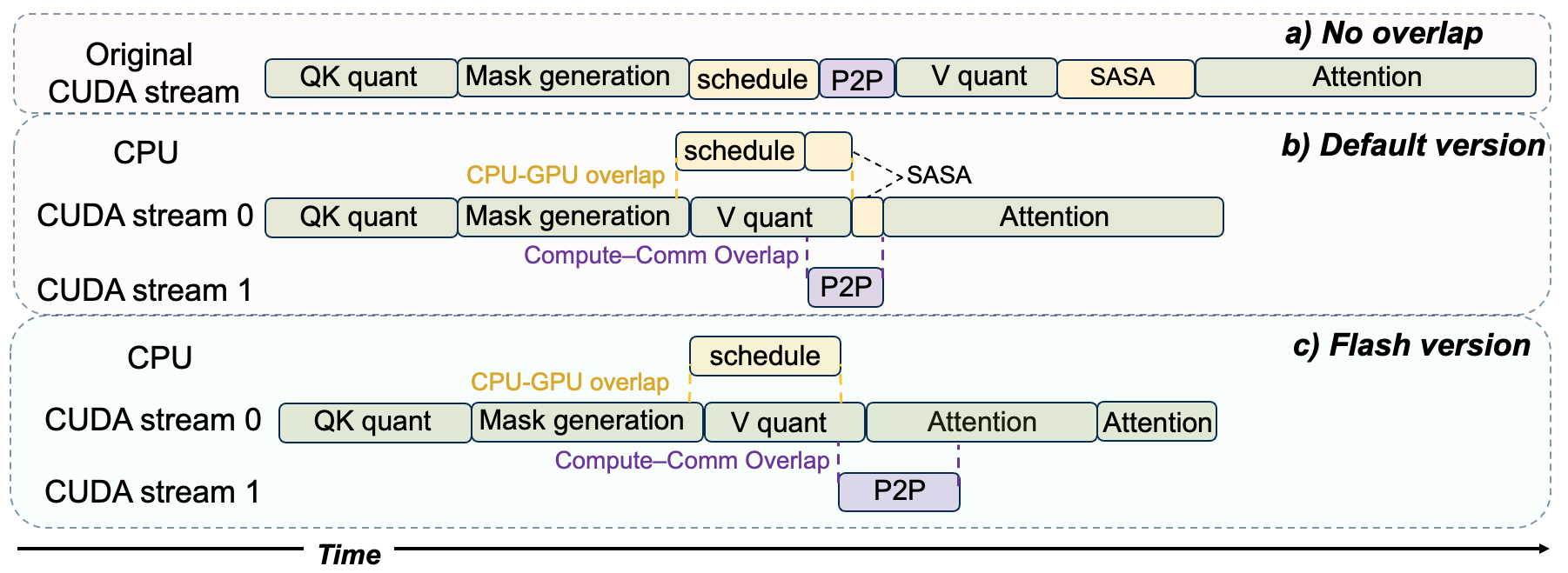}
  \caption{Overlapped execution schedule.}
  \label{fig:overlap}
\end{figure}

Scheduling search, P2P migration, and augmentation are issued non-blockingly and synchronized only at true data dependencies.
Beyond asynchronous overlap, Overlap/Opt. removes unnecessary CPU--GPU synchronization and streamlines runtime management and kernel launches, preserving the algorithmic decisions of \rlb{} and \sasa{} while reducing visible cost.
Table~\ref{tab:overhead} shows that the added runtime components expose only $1.87\%$ visible overhead after Overlap/Opt.
The complete stack reduces attention latency from $41.33$ ms to $37.54$ ms without changing the load distribution.
Together, \rlb{}, \sasa{}, and Overlap/Opt. repair imbalance, reinvest residual slack into mask fidelity, and minimize visible runtime cost.

\section{Experiments}
\label{sec:experiments}

\subsection{Experimental Setup}
\textbf{Models and workloads.}
We evaluate \method{} on three video DiT workloads: Wan2.2 I2V 14B, Wan2.2 Animate 14B, and Wan2.1 T2V 14B.
All workloads use the LightX2V 4-step distilled LoRA configuration~\citep{lightx2v}, representing the few-step inference regime targeted by this work.
Unless otherwise stated, each experiment generates 720p videos with 81 frames, corresponding to 21 latent frames and 3,600 tokens per latent frame.
For T2V and I2V, we use the complete VBench~\citep{huang2024vbench,huang2025vbench++} evaluation sets, containing 946 and 1,118 test cases, respectively.
For Animate, we use an internal evaluation set containing 20 test cases.
All compared methods use identical evaluation inputs and inference configurations.

\textbf{Metrics and hardware.}
We use VBench to measure overall video generation quality following the official evaluation protocol.
To quantify how closely sparse attention preserves the dense output, we also report PSNR, SSIM, LPIPS, and CLIP similarity (CLIP-Sim) against the FlashAttention~\citep{dao2022flashattention} output under the same prompt and seed.
Efficiency is measured by average DiT latency under 8-GPU Ulysses sequence parallelism and by the corresponding speedup over FlashAttention. 
DiT latency measures complete denoising-transformer execution over all sampling steps, excluding text encoding and VAE encoding/decoding, and accounts for the dominant portion of end-to-end generation latency.
Unless otherwise stated, attention latency spans sequence-to-head All-to-All through reverse head-to-sequence All-to-All, including mask routing, scheduling, and sparse-attention execution.
All experiments are conducted on a server with 8$\times$NVIDIA H20 GPUs, connected via NVLink, using CUDA~13.1.

\textbf{Baselines and variants.}
FlashAttention serves as the dense baseline.
We compare against representative training-free sparse-attention methods for video generation, including SpargeAttention~\citep{zhang2025spargeattn} with Top-$k$ or Top-$p$ selection, Jenga~\cite{zhang2026training}, and SVG2~\citep{yang2026sparse}, as well as dense quantized SageAttention~\citep{zhang2025sageattention}.
For load balancing, we compare against no load balancing and db-SP~\citep{chen2025db} on both SpargeAttention and \method{}-Base frontends.
\method{}-Base denotes our adaptive sparse-routing frontend without \rlb{} or \sasa{}, and the full \method{} is \method{}-Base plus Runtime Load Balancing and Slack-Aware Sparse Augmentation.
We evaluate \method{} under multiple Top-$p$ sparsity settings and use Top-$p=0.95$ and Top-$p=0.90$ as the main quality-oriented and speed-oriented operating points.
All sparse-attention methods execute dense attention for the first $25\%$ of denoising steps.
For SVG2, we use its original FlashInfer-based dynamic-block backend and adjust the $k$-means initialization count for the 4-step setting.

\textbf{Implementation details.}
The piecewise thresholds are set to $k_1{=}1.05$ and $k_2{=}1.10$: \sasa{} alone is enabled when $k_1\le\rho<k_2$, and \rlb{}+\sasa{} when $\rho\ge k_2$.
\rlb{} migrates at most one local head per rank (roughly $20\%$).
\sasa{} uses augmentation coefficient $n_1{=}0.8$ and trigger threshold $n_2{=}0.07$.

\textbf{Evaluation roadmap.}
The experiments proceed in three parts.
We first report end-to-end quality--efficiency results across I2V, Animate, and T2V workloads.
We then isolate the effects of \rlb{}, \sasa{}, and overlap through end-to-end and attention-runtime ablations, including scaling and imbalance studies.
Finally, we decompose the standalone and visible runtime overheads of the added mechanisms, connecting the measurements back to the joint gain model $G=K-P-C-A$.

\subsection{Overall Performance}
\begin{table}[tbp]
\centering
\caption{I2V and T2V end-to-end results.}
\label{tab:i2v-t2v-main}
\small
\adjustbox{max width=\linewidth}{
\begin{tabular}{@{}llccccccc@{}}
\toprule
Method & Config & VBench$\uparrow$ & PSNR$\uparrow$ & SSIM$\uparrow$ & LPIPS$\downarrow$ & CLIP-Sim$\uparrow$ & DiT Latency$\downarrow$ & Speedup$\uparrow$ \\
\midrule
\multicolumn{9}{@{}l}{\textit{Wan2.2-14B-I2V 4 steps}} \\
FlashAttention & Dense & 88.7\% & -- & -- & -- & -- & 38.59s & 1.00$\times$ \\
SageAttention & Dense & 88.7\% & 25.414 & 0.8420 & 0.0868 & 0.9922 & 21.20s & 1.82$\times$ \\
SVG2 & Top-$p$=0.80 & 88.7\% & 22.967 & 0.7947 & 0.1160 & 0.9889 & 29.96s & 1.29$\times$ \\
SpargeAttention & Top-$p$=0.95 & 88.3\% & 22.400 & 0.7728 & 0.1261 & 0.9880 & 19.92s & 1.94$\times$ \\
SpargeAttention & Top-$p$=0.90 & 88.2\% & 21.477 & 0.7425 & 0.1472 & 0.9857 & 19.23s & 2.01$\times$ \\
SpargeAttention + db-SP & Top-$p$=0.90 & 88.2\% & 21.477 & 0.7425 & 0.1472 & 0.9857 & 19.09s & 2.02$\times$ \\
SpargeAttention + \rlb{} + \sasa{} & Top-$p$=0.90 & 88.3\% & 21.807 & 0.7513 & 0.1381 & 0.9865 & 18.84s & 2.05$\times$ \\
\rowcolor{tabgray}
\textbf{\method{}} & Top-$p$=0.95 & \textbf{88.8\%} & \textbf{23.801} & \textbf{0.8092} & \textbf{0.1045} & \textbf{0.9906} & 19.10s & 2.02$\times$ \\
\rowcolor{tabgray}
\textbf{\method{}} & Top-$p$=0.90 & \textbf{88.8\%} & 23.473 & 0.8024 & 0.1091 & 0.9903 & \textbf{18.30s} & \textbf{2.11$\times$} \\
\midrule
\multicolumn{9}{@{}l}{\textit{Wan2.1-14B-T2V 4 steps}} \\
FlashAttention & Dense & 81.3\% & -- & -- & -- & -- & 34.70s & 1.00$\times$ \\
\rowcolor{tabgray}
\textbf{\method{}} & Top-$p$=0.90 & \textbf{81.6\%} & \textbf{19.585} & \textbf{0.7454} & \textbf{0.1411} & \textbf{0.9849} & \textbf{14.96s} & \textbf{2.32$\times$} \\
SpargeAttention & Top-$p$=0.90 & 80.9\% & 17.161 & 0.6438 & 0.2119 & 0.9746 & 15.69s & 2.21$\times$ \\
SpargeAttention + db-SP & Top-$p$=0.90 & 80.9\% & 17.161 & 0.6438 & 0.2119 & 0.9746 & 15.47s & 2.24$\times$ \\
SpargeAttention + \rlb{} + \sasa{} & Top-$p$=0.90 & 81.0\% & 18.571 & 0.6999 & 0.1700 & 0.9812 & 15.31s & 2.27$\times$ \\
\bottomrule
\end{tabular}
}
\end{table}

\begin{table}[tbp]
\centering
\caption{Wan2.2 Animate end-to-end results.}
\label{tab:animate-main}
\small
\adjustbox{max width=\linewidth}{
\begin{tabular}{@{}lcccccc@{}}
\toprule
Method & PSNR$\uparrow$ & SSIM$\uparrow$ & LPIPS$\downarrow$ & CLIP-Sim$\uparrow$ & DiT Latency$\downarrow$ & Speedup$\uparrow$ \\
\midrule
FlashAttention & -- & -- & -- & -- & 36.95s & 1.00$\times$ \\
SageAttention & 22.259 & 0.8386 & 0.1131 & 0.9871 & 17.96s & 2.06$\times$ \\
\method{}-Base (Top-$p$=0.95) & 22.182 & 0.8343 & 0.1239 & 0.9848 & 16.40s & 2.25$\times$ \\
\rowcolor{tabgray}
\textbf{\method{} (Top-$p$=0.95)} & \textbf{22.640} & \textbf{0.8488} & \textbf{0.1066} & \textbf{0.9873} & 15.21s & 2.43$\times$ \\
\method{}-Base + db-SP & 22.182 & 0.8343 & 0.1239 & 0.9848 & 15.51s & 2.38$\times$ \\
Jenga & 21.534 & 0.8256 & 0.1314 & 0.9831 & 22.24s & 1.66$\times$ \\
\method{} (Top-$p$=0.90) & 21.512 & 0.8257 & 0.1291 & 0.9851 & \textbf{14.79s} & \textbf{2.50$\times$} \\
SpargeAttention (Top-$k$=0.60) & 21.188 & 0.8169 & 0.1400 & 0.9824 & 15.71s & 2.35$\times$ \\
SVG2 (Top-$p$=0.95) & 21.002 & 0.8167 & 0.1389 & 0.9817 & 29.14s & 1.27$\times$ \\
\bottomrule
\end{tabular}
}
\end{table}

Tables~\ref{tab:i2v-t2v-main} and~\ref{tab:animate-main} show that \method{} consistently improves the end-to-end quality--efficiency trade-off across I2V, Animate, and T2V workloads.
More specifically, \method{} exhibits both ``same speed, higher quality'' and ``same quality, higher speed'' behavior.
On I2V, \method{} at Top-$p$=0.95 has a DiT latency close to SpargeAttention at Top-$p$=0.95 (19.10s vs. 19.92s), but substantially improves PSNR/SSIM/LPIPS/CLIP-Sim (23.801/0.8092/0.1045/0.9906 vs. 22.400/0.7728/0.1261/0.9880).
At Top-$p$=0.90, \method{} reduces DiT latency from 19.23s to 18.30s compared with SpargeAttention while also improving VBench (88.8\% vs. 88.2\%) and all fidelity metrics.
The Animate and T2V results follow the same trend: \method{} preserves higher fidelity at comparable or lower latency, and provides higher acceleration when quality is comparable.
Together, these results place \method{} on a stronger Pareto frontier than existing training-free sparse-attention baselines.

The controlled SpargeAttention rows further isolate the effect of runtime scheduling.
Under the same Top-$p$ sparse-routing frontend, db-SP provides only limited speedup over no load balancing, whereas \rlb{}+\sasa{} further reduces DiT latency and improves fidelity.
For example, on I2V, SpargeAttention+\rlb{}+\sasa{} reduces DiT latency from 19.09s to 18.84s over SpargeAttention+db-SP and improves PSNR from 21.477 to 21.807.
On T2V, it reduces DiT latency from 15.47s to 15.31s and improves PSNR/SSIM/LPIPS/CLIP-Sim from 17.161/0.6438/0.2119/0.9746 to 18.571/0.6999/0.1700/0.9812.
This indicates that current-workload repair and residual-slack reuse are more suitable than history-based layout reuse for few-step video DiT inference, and that \rlb{}+\sasa{} can serve as plug-and-play runtime components for Top-$p$-based sparse attention.

\subsection{Ablation Studies}

We ablate \rlb{}, \sasa{}, and overlap along two axes.
Table~\ref{tab:e2e-ablation} measures their end-to-end effect on Wan2.2 Animate, while Table~\ref{tab:kernel-ablation} isolates the attention-kernel behavior on Wan2.2 I2V.
Together, these experiments connect the quality--latency trade-off to the underlying load imbalance and critical-path cost.

\begin{table}[tbp]
\centering
\caption{End-to-end ablation on Wan2.2 Animate.
\rlb{} changes only the head-to-rank placement, while \sasa{} spends residual slack on additional high-value sparse blocks.}
\label{tab:e2e-ablation}
\small
\adjustbox{max width=\linewidth}{
\begin{tabular}{@{}lcccccc@{}}
\toprule
Configuration & PSNR$\uparrow$ & SSIM$\uparrow$ & LPIPS$\downarrow$ & CLIP-Sim$\uparrow$ & DiT Latency$\downarrow$ & Speedup$\uparrow$ \\
\midrule
FlashAttention & -- & -- & -- & -- & 36.95s & 1.00$\times$ \\
\method{}-Base (Top-$p$=0.95) & 22.182 & 0.8343 & 0.1239 & 0.9848 & 16.40s & 2.25$\times$ \\
\quad + \rlb{} & 22.182 & 0.8343 & 0.1239 & 0.9848 & \textbf{15.18s} & \textbf{2.43$\times$} \\
\rowcolor{tabgray}
\quad + \rlb{} + \sasa{} & \textbf{22.640} & \textbf{0.8488} & \textbf{0.1066} & \textbf{0.9873} & 15.21s & \textbf{2.43$\times$} \\
\method{}-Base + db-SP & 22.182 & 0.8343 & 0.1239 & 0.9848 & 15.51s & 2.38$\times$ \\
\bottomrule
\end{tabular}
}
\end{table}

The end-to-end ablation separates the roles of the two runtime mechanisms.
Adding \rlb{} leaves PSNR, SSIM, LPIPS, and CLIP-Sim unchanged from \method{}-Base (22.182/0.8343/0.1239/0.9848), confirming that head migration only changes where heads execute and does not alter the selected QK blocks.
Despite this quality-preserving behavior, it reduces DiT latency from 16.40s to 15.18s (7.4\%) and improves the speedup from 2.25$\times$ to 2.43$\times$.
At identical sparse-routing quality, \rlb{} is also 2.2\% faster than db-SP (15.18s vs. 15.51s), supporting the design choice of repairing the current realized workload rather than reusing a historical layout.

\sasa{} provides the complementary effect: it improves sparse-mask coverage with negligible latency cost.
On top of \rlb{}, \sasa{} improves PSNR by 0.458\,dB (22.182$\rightarrow$22.640), SSIM by 1.7\% (0.8343$\rightarrow$0.8488), and CLIP-Sim from 0.9848 to 0.9873, while reducing LPIPS by 13.9\% (0.1239$\rightarrow$0.1066).
The DiT latency changes only from 15.18s to 15.21s, indicating that the augmentation budget remains within residual slack and does not extend the critical path.

\begin{table}[tbp]
\centering
\caption{Attention-runtime ablation on Wan2.2 I2V.}
\label{tab:kernel-ablation}
\small
\adjustbox{max width=\linewidth}{
\begin{tabular}{@{}lccc@{}}
\toprule
Method & Attention Latency$\downarrow$ & Final Imbalance $\rho\downarrow$ & Speedup vs. Flash$\uparrow$ \\
\midrule
FlashAttention & 165.60 ms & 1.00 & 1.00$\times$ \\
\method{}-Base (Top-$p$=0.95) & 45.91 ms & 1.34 & 3.61$\times$ \\
\quad + \rlb{} & 41.32 ms & 1.08 & 4.01$\times$ \\
\quad + \rlb{} + \sasa{} & 41.33 ms & 1.01 & 4.00$\times$ \\
\rowcolor{tabgray}
\quad + \rlb{} + \sasa{} + Overlap/Opt. & \textbf{37.54 ms} & \textbf{1.01} & \textbf{4.41$\times$} \\
\method{}-Base + db-SP & 44.17 ms & 1.22 & 3.75$\times$ \\
\bottomrule
\end{tabular}
}
\end{table}

\begin{figure}[tbp]
  \centering
  \includegraphics[width=0.78\linewidth]{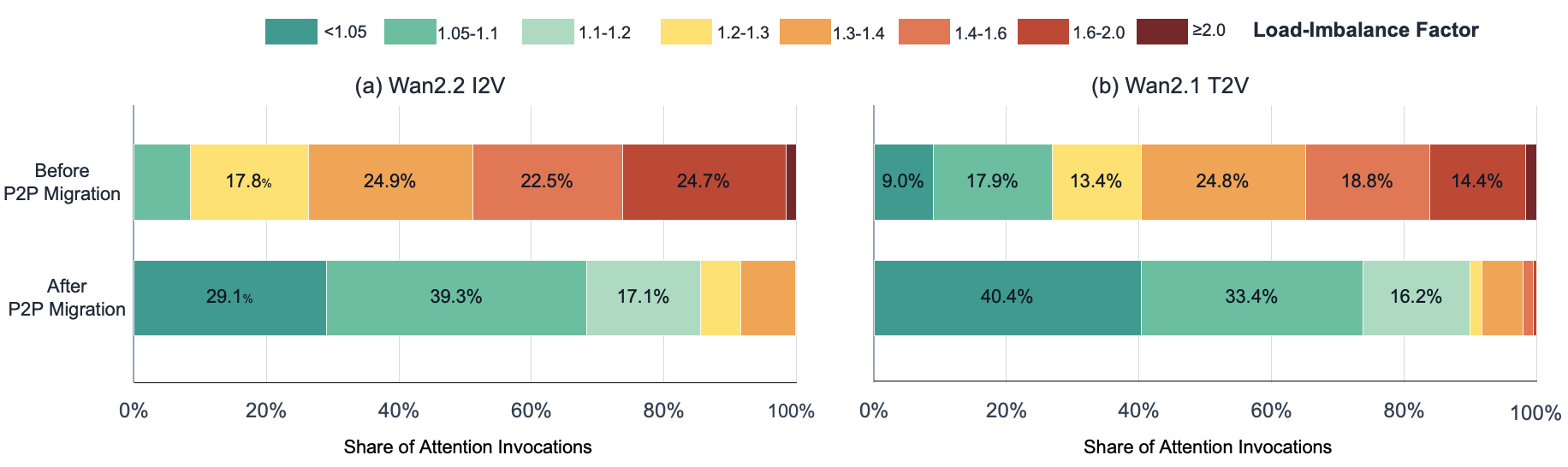}
  \caption{Distribution of the load-imbalance factor across attention invocations before and after RLB.}
  \label{fig:lb-distribution}
\end{figure}

\begin{figure}[tbp]
  \centering
  \includegraphics[width=0.82\linewidth]{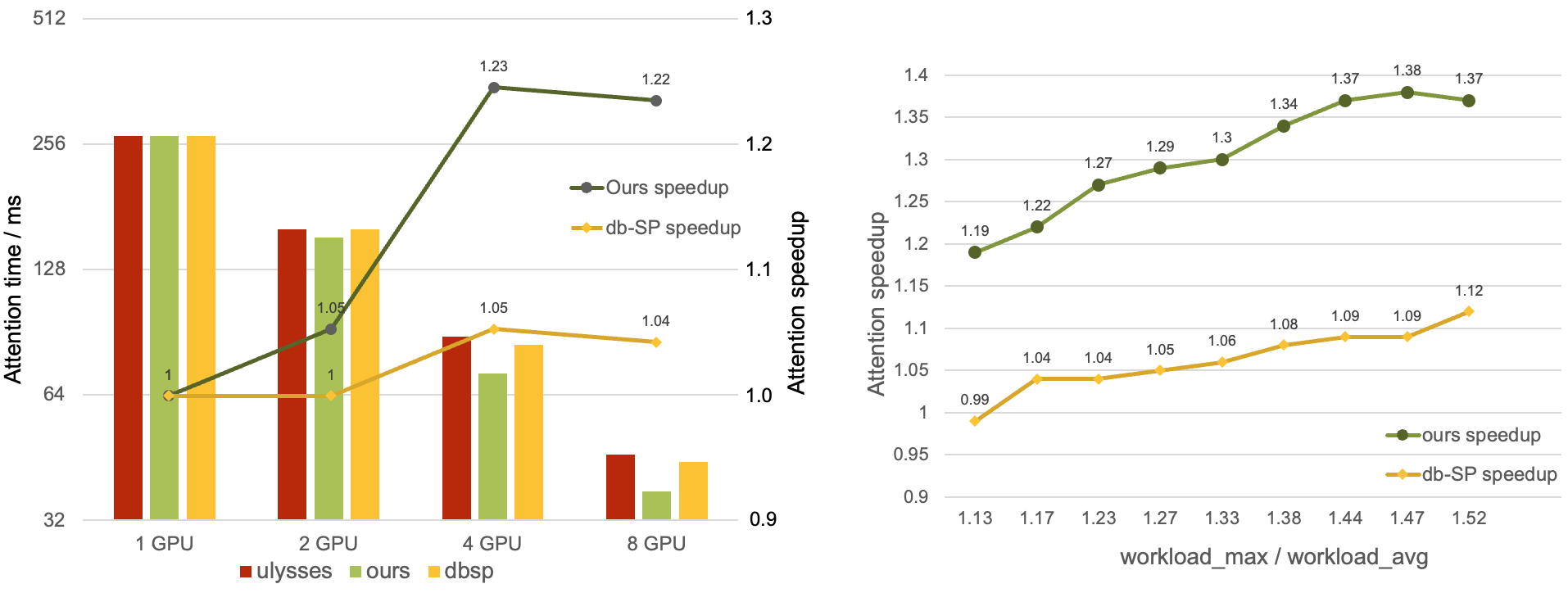}
  \caption{Runtime Load Balancing remains faster than db-SP across the evaluated GPU counts and provides larger gains when the initial imbalance is higher.}
  \label{fig:scaling-imbalance}
\end{figure}

Table~\ref{tab:kernel-ablation} provides mechanism-level evidence for the three design goals in Sections~\ref{sec:rlb}--\ref{sec:overlap}: reducing critical-path workload, filling residual slack with useful blocks, and hiding the visible cost of runtime repair.
Adaptive sparse routing alone is fast but creates rank-level skew: \method{}-Base reaches 3.61$\times$ attention speedup over FlashAttention, yet its final imbalance rises to $\rho=1.34$.
\rlb{} reduces this imbalance to 1.08 and lowers attention latency from 45.91 ms to 41.32 ms, increasing the speedup to 4.01$\times$.
This matches the gain model $K=c(\max_r L_r-\max_r L_r^{\mathrm{new}})$: reducing the maximum realized rank load directly shortens the critical path.
Compared with db-SP, \rlb{} obtains both lower imbalance (1.08 vs. 1.22) and lower latency (41.32 ms vs. 44.17 ms), showing that per-invocation repair better matches the current few-step video DiT workload than a reused historical layout.

The effect of \sasa{} should be interpreted as slack filling rather than another conventional load-balancing step.
After \rlb{}, \sasa{} brings the final effective imbalance from 1.08 to 1.01 because non-critical ranks spend their residual slack on additional high-value sparse blocks.
The attention latency remains essentially unchanged (41.32 ms to 41.33 ms), which is consistent with the constraint $B_r\le \Delta L_r$: augmentation should improve useful work coverage without creating a new straggler.
Finally, Overlap/Opt.\ reduces attention latency from $41.33$ ms to $37.54$ ms and improves the speedup from $4.00\times$ to $4.41\times$.
Because $\rho$ remains $1.01$, this gain comes from execution-level optimization rather than additional load balancing or sparse augmentation.

Figure~\ref{fig:scaling-imbalance} further characterizes when runtime load balancing is most beneficial.
When the initial imbalance is low, there is little straggler time to remove, so the relative gain is modest; this explains the smaller benefit in the 2-GPU setting.
As the initial skew increases, \rlb{} yields larger gains because it directly targets the current critical path.
Across the evaluated GPU counts and imbalance degrees, \rlb{} consistently outperforms db-SP, whose historical layout can become stale under few-step generation and dynamic sparse routing.

\subsection{Overhead Analysis}

\begin{table}[tbp]
\centering
\caption{Incremental overhead of the runtime components added by \method{} on Wan2.2 I2V. Standalone cost measures each component in isolation, while visible cost is the portion remaining after Overlap/Opt.}
\label{tab:overhead}
\small
\adjustbox{max width=\linewidth}{
\begin{tabular}{@{}lcccc@{}}
\toprule
Component & Standalone Added Cost & Ratio w.r.t. Attention & Visible Added Cost after Overlap/Opt. & Visible Ratio \\
\midrule
Density exchange & $\sim$0.3 ms & 0.72\% & $\sim$0.1 ms & 0.27\% \\
Balance plan search & $\sim$0.5 ms & 1.21\% & $\sim$0.1 ms & 0.27\% \\
P2P head migration & $\sim$0.6 ms & 1.45\% & $\sim$0.1 ms & 0.27\% \\
Slack-aware augmentation & $\sim$0.5 ms & 1.21\% & $\sim$0.2 ms & 0.53\% \\
Head-order restoration & $\sim$0.1 ms & 0.24\% & $\sim$0.1 ms & 0.27\% \\
Other overhead (metadata, kernel launch) & $\sim$0.6 ms & 1.45\% & $\sim$0.1 ms & 0.27\% \\
\midrule
\textbf{Total overhead} & \textbf{$\sim$2.6 ms} & \textbf{6.28\%} & \textbf{$\sim$0.7 ms} & \textbf{1.87\%} \\
\textbf{Reference attention latency} & \textbf{41.4 ms} & \textbf{100\%} & \textbf{37.5 ms} & \textbf{100\%} \\
\bottomrule
\end{tabular}
}
\end{table}

Table~\ref{tab:overhead} decomposes the incremental overhead introduced by the added runtime mechanisms under the Top-$p$=0.95 + \rlb{} + \sasa{} + Overlap/Opt. configuration.
The standalone column measures the serialized cost of each added component, while the visible column measures the portion remaining on the critical path after Overlap/Opt.
To connect these measurements to the joint gain model $G=K-P-C-A$, density exchange and P2P head migration correspond to communication cost $P$, balance-plan search and head-order restoration correspond to scheduling cost $C$, and slack-aware augmentation corresponds to augmentation cost $A$.
Importantly, Table~\ref{tab:overhead} decomposes only these incremental costs; it does not fully explain the reduction from 41.33 ms to 37.54 ms in Table~\ref{tab:kernel-ablation}, because Overlap/Opt. also removes synchronization and execution inefficiencies from the underlying attention path.

The two overlap paths in Section~\ref{sec:overlap} reduce the visible portion of these added costs.
CPU--GPU overlap hides balance-plan search and budget computation inside the GPU-side V-quantization window,
while computation--communication overlap hides density exchange and P2P migration behind non-migrated-head computation and augmentation work.
\sasa{} retains the largest visible added cost ($\sim$0.2 ms, 0.53\%) because part of its mask update and augmentation remains on the critical path;
nevertheless, this residual cost remains small under the slack constraint.

After Overlap/Opt., the added runtime components expose only $\sim$0.7 ms of visible critical-path overhead, or 1.87\% of the final 37.5 ms attention latency.
By reducing the visible portion of $P+C+A$ while preserving the workload distribution, Overlap/Opt.\ further increases the net gain $G$.

\section{Conclusion and Limitations}
\label{sec:conclusion}

\paragraph{Conclusion.}
We presented \method{}, a training-free sparse-attention system for video DiT inference that improves the distributed execution efficiency of dynamic sparse routing under multi-GPU sequence parallelism.
Adaptive Top-$p$ sparse attention can allocate compute budgets according to the information demand of each attention head and improve mask fidelity, but it also creates significant head-level workload heterogeneity that becomes rank-level stragglers in multi-GPU execution.
To address this tension, \method{} performs lightweight P2P head migration after the current sparse mask is materialized, using Runtime Load Balancing to repair the critical path of the current attention phase.
On top of this repair, Slack-Aware Sparse Augmentation converts the residual slack on non-critical ranks into additional high-value sparse blocks, improving sparse-mask coverage without significantly increasing the global attention latency.
Through CPU--GPU overlap and computation--communication overlap, \method{} keeps the visible overhead of scheduling, migration, and augmentation at a low level.
Experiments on Wan2.2 I2V, Wan2.2 Animate, and Wan2.1 T2V show that \method{} simultaneously improves load balance, attention latency, DiT latency, and video quality metrics, validating the effectiveness of \rlb{} and \sasa{}.

\paragraph{Limitations.}
\method{} primarily targets long spatiotemporal sequence video generation.
For short sequences or workloads with weak attention sparsity, such as many image-generation tasks, both the sparse-attention saving and the load-balancing or slack-reuse benefit decrease accordingly.
The benefit of \method{} also depends on the communication-to-computation ratio of the hardware.
On H20-class servers with sufficient communication and memory bandwidth, lightweight P2P migration and overlap can effectively hide the runtime overhead.
On PCIe-connected devices with stronger compute but weaker inter-GPU communication, the gain from head migration may be limited, and the migration budget, SASA trigger threshold, and overlap strategy may need to be recalibrated.
Furthermore, this work validates \method{} mainly under Ulysses-style sequence parallelism and few-step video DiT inference.
Future work can explore its applicability under Ring Attention, USP, more diverse hardware topologies, and training-time sparse attention.

\bibliographystyle{plainnat}
\bibliography{references}

\appendix

\end{document}